\documentclass[nohyperref]{article}

\usepackage{microtype}
\usepackage{graphicx}
\usepackage{subfigure}
\usepackage{booktabs} 

\usepackage[accepted]{icml2023}

\usepackage{amsmath}
\usepackage{amssymb}
\usepackage{mathtools}
\usepackage{amsthm}

\usepackage[capitalize,noabbrev]{cleveref}

\theoremstyle{plain}

\theoremstyle{definition}

\theoremstyle{remark}

\usepackage[textsize=tiny]{todonotes}

\RequirePackage{color}

\icmltitlerunning{Studying Generalization on Memory-Based Methods in Continual Learning}

\begin{document}

\twocolumn[
\icmltitle{Studying Generalization on Memory-Based Methods in Continual Learning}



\icmlsetsymbol{equal}{*}

\begin{icmlauthorlist}
\icmlauthor{Felipe del Rio}{puc}
\icmlauthor{Julio Hurtado}{unipi,cenia}
\icmlauthor{Cristian Buc}{cenia}
\icmlauthor{Alvaro Soto}{puc,cenia}
\icmlauthor{Vincenzo Lomonaco}{unipi}
\end{icmlauthorlist}

\icmlaffiliation{puc}{Department of Computer Science, Pontificia Universidad Católica de Chile, Santiago, Chile}
\icmlaffiliation{unipi}{Università di Pisa, Pisa, Italy}
\icmlaffiliation{cenia}{Centro Nacional de Inteligencia Artificial (CENIA), Santiago, Chile}

\icmlcorrespondingauthor{Felipe del Rio}{fidelrio@uc.cl}

\icmlkeywords{Machine Learning, ICML}

\vskip 0.3in
]



\printAffiliationsAndNotice{}  

\begin{abstract}
One of the objectives of Continual Learning is to learn new concepts continually over a stream of experiences and at the same time avoid catastrophic forgetting. To mitigate complete knowledge overwriting, memory-based methods store a percentage of previous data distributions to be used during training. Although these methods produce good results, few studies have tested their out-of-distribution generalization properties, as well as whether these methods overfit the replay memory. In this work, we show that although these methods can help in traditional in-distribution generalization, they can strongly impair out-of-distribution generalization by learning spurious features and correlations. Using a controlled environment, the Synbol benchmark generator \cite{lacoste2020synbols}, we demonstrate that this lack of out-of-distribution generalization mainly occurs in the linear classifier.
\end{abstract}

\vspace{-6mm}

\section{Introduction}

\vspace{-1mm}

Continual Learning (CL) aims to develop models and training procedures capable of learning continuously through a stream of data \cite{delange2021continual}. As opposed to the well-studied static setting of feeding the model with independent and identically distributed (IID) data, in CL, each experience has its distribution with a possible drift among tasks.

Given this distribution drift, one of the main challenges of CL is catastrophic forgetting \cite{mccloskey1989catastrophic}. The latter refers to the process by which a model forgets to solve previously learned tasks when new experiences come in. In this context, replay-based methods provide a powerful and straightforward tool to counter catastrophic forgetting by storing and revisiting a subset of samples from previously learned tasks. These methods have achieved state-of-the-art results in a wide array of continual learning scenarios and benchmarks \cite{chaudhry2019tiny, buzzega2020dark}.

Despite successful results, previous works have argued that memory-based methods are prone to overfitting \cite{lopez2017gradient, verwimp2021rehearsal}. By only storing a subset of previous distributions, the model only reinforces concepts and ideas that are present in the buffer, depending on how much previous distributions are represented. To reinforce useful concepts, the buffer should accurately represent the whole training distribution. However, if the buffer represents only a small percentage of the training distribution, it will start learning spurious correlations and will lose its generalization capabilities. 

We argue that compositionality is a critical factor for CL. However, spurious correlations in the data can lead the model to learn incorrect compositions of specific concepts, thus impairing generalization. In this paper, we show that, even if a model can learn to identify useful concepts to make a proper classification, the classifier will learn shortcuts that hurt out-of-distribution generalization (OOD); shortcuts that help increase performance in the IID dataset and are amplified by memory-based methods. However, as a result, they increase the generalization gap between IID and OOD examples.

In this paper, we develop a controlled setting that tests out-of-distribution generalization beyond the training distribution. We evaluate a basic CNN model on a set of examples that depart from the training distribution by including unseen combinations of latent and target variables. And we show that replay falters in this setting, giving further evidence that replay-based methods have a toll on generalization capabilities not seen on traditional machine learning benchmarks that test only the IID test set. Here we propose an approach to test how OOD and spurious correlations affect memory-based methods and hope our results influence future studies to focus on improving the performance of CL in OOD data.

\vspace{-2mm}

\section{Related Work}

\vspace{-1mm}


Memory-based methods address catastrophic forgetting by incorporating data from previous tasks into the training process \cite{lopez2017gradient, buzzega2020dark}. Most methods save samples from previous experiences to be used in the current task \cite{rebuffi2017icarl, chaudhry2019tiny}, hoping these can be a good representation of past distributions and maintain performance. Previous research suggests that rehearsal would result in over-fitting, affecting generalization \cite{lopez2017gradient, verwimp2021rehearsal}. However, other researchers suggest the opposite \cite{chaudhry2019tiny}, making this issue still an open question \cite{peng2023ideal}.

One way to tackle the problem of generalization in machine learning is compositionality \cite{hupkes2020compositionality,lake2019compositional}.
Compositional representations refer to decomposing concepts in their sub-parts. Learning these representations is useful, as these can be recombined to create novel concepts or make sense of new experiences \cite{hurtado2021optimizing, mendez2022reuse, veniat2021efficient}. Previous works have used compositionality to this effect \cite{loula2018rearranging,lake2019compositional,chen2020compositional,akyureklearning}.
In CL, \citealp{ostapenko2021cl-local-module-composition} uses explicit compositions of neural modules as a way to reuse knowledge from previous tasks to solve new ones and show this approach increases the model's generalization capabilities.

\vspace{-2mm}

\section{Experimental Setting}

\vspace{-1mm}

The common practice in CL dictates that, for each experience, the distribution of training and test sets follows a similar distribution, and changes in the distributions take place with new experiences. However, in an uncontrolled setting, changes in the distribution of a known experience can occur as it is almost impossible to completely represent the full distribution with a subset of samples, e.g., sample selection bias \cite{quinonero2008dataset}. To tackle this issue, we aim for testing the generalization capabilities of a model in OOD set together with the IID test set. This new set should present similar characteristics to the training set, but with a systematically different distribution, e.g., leaving out some combination of concepts.

In CL, we assume that each experience $t$ of the sequence follows a distribution $P_t(y_i | x_i, z_i)$, where $y_i$ is the label, $x_i$ is the input, and $z_i$ are a group of characteristics presented in the input. In a classification task, the objective is to minimize the function $\mathcal{L}_t (f_{\Theta}(x_i^t), y_i^t)$ for every experience of the sequence, meaning that the model $f$ must learn parameters $\Theta$ that find relevant characteristics from $z_{i}$ that generalize to solve the current and future task using only information from the input. However, as it has been shown in previous studies \cite{geirhos2020shortcut-learning,ming2022impact}, it is common that the model uses shortcuts and learns spurious features that help the model solve the task without generalizing to OOD samples. To test this, we follow a strategy used in systematic and compositional generalization research \cite{lake18scan,ruis2020gscan,kim2020cogs,keysers2020measuring}, and propose the creation of an OOD test set to quantify the ability of the model to generalize to examples that drift from the training distribution.


To test OOD generalization, we must know which attributes $z$ are useful for solving a particular task. Ideally, a model should be able to correctly identify relevant features $z_g$ and irrelevant features $z_b$. In this paper, we will assume that a model with proper OOD generalization properties can correctly identify those relevant features and combine them to solve the task. In contrast, a model relying on spurious correlations is one that can correctly encode relevant features but incorrectly extracts this information or uses irrelevant features for solving the task. 

We create a dataset where we control every characteristic $z$ that generates an image. We identify one of these characteristics as the label $y$, a group of relevant features to solve the task $z_g$, and the rest as irrelevant features. In order for the model to be effective, it must identify features $z_g$ and combine them to correctly identify $y$. We expect that even with missing combinations of $y$, $z_{g}$ from the training set, a robust model that is able to identify $y$ and $z_{g}$ independently can extrapolate its knowledge correctly to solve the task despite the absent combinations. 

\begin{figure}
    \centering
    \includegraphics[width=0.85\linewidth]{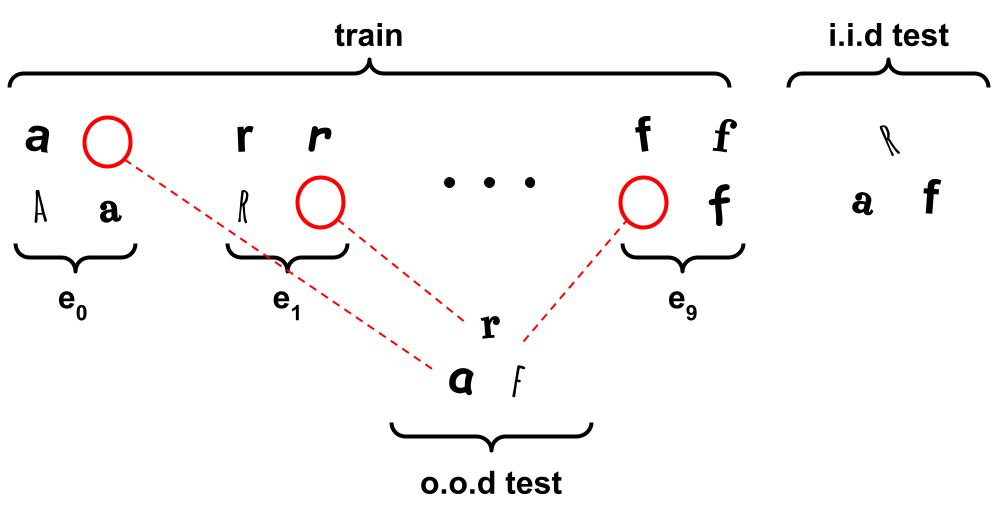}
    \vspace{-3mm}
    \caption{Illustration of the generated continual learning task by dividing the train set by the selected latent variable, the character present.
    The IID test set is sampled from the union of all training distributions.
    And the OOD test set uses combinations of character-font not present during training to test for non-spurious generalization. It is important to note that even if a combination character-font is not present in the training set, every all fonts are present in every task.
\vspace{-3mm}}
    \label{fig:continual-learning-scenario}
    \vspace{-5mm}
\end{figure}

For the CL scenario, we create the sequence following a domain incremental setup \cite{van2019three}, each experience contains every class and a different domain distribution. Each experience will have a disjoint group of features $z_g$ present. For every experience $t, t_1, t_2 \in \{1,2,...,E\}$, $z_t=\{\hat{z}_0,\hat{z}_1,...,\hat{z}_{|z_t|}\}$, where $z_{t_1} \cap z_{t_2} = \emptyset,  \forall {t_1} \neq {t_2}$. We create a test set that follows the same distribution $P_t(y_i | x_i, z_i)$ of each experience. In addition to the test set, we built a second set, which we call a generalization set, to test the OOD generalization capabilities of the model. This is achieved by holding out a number of combinations of features-label tuples $(z_j,y_j) \in G_t$ from being given to the model during training $P_t(z_j,y_j) = 0, \forall t \in E$. An example of the training and testing scenario generated can be seen in Fig. \ref{fig:continual-learning-scenario}.

\vspace{-2mm}

\section{Experimental Set-up}

\subsection{Benchmark}

\begin{figure}
    \centering
    \includegraphics[width=0.9\linewidth]{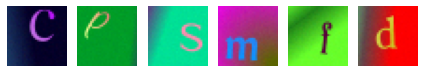}
    \vspace{-3mm}
    \caption{Examples of character images generated by the Synbols generator, and used in our training set. Our task uses these images to predict the font present on them.
\vspace{-3mm}}
    \label{fig:synbols-samples}
    \vspace{-3mm}
\end{figure}

We leverage the Synbol benchmark framework \cite{lacoste2020synbols} to quickly create a synthetic dataset composed of images of different characters with various sizes, positions, fonts, colors, and backgrounds. Figure \ref{fig:synbols-samples} shows examples of the characters generated.
This benchmark allows us to access the relevant features or latent variables $z$ used to generate each image.
Thus, allowing us the flexibility to create a task like the one described in the previous section.

We use the font prediction task for our experiments, i.e., the font is the task-relevant factor.
For the creation of experiences, we use English non-diacritic characters as the feature $z$ to partition the dataset, this choice reduces the similarity between domains. 
We sample $10,000$ images in total, made from $10$ fonts and $10$ characters.
These images are used to create the splits sets, namely train and, IID and OOD test.

We create five different CL scenarios by dividing the dataset into different numbers of tasks $T$, namely $T \in \{1, 2, 4, 5, 10\}$. 
When $T=1$, the scenario consists of the static setting and we use it for comparison.
Each task $t$ is created by assigning a number of characters to it and assigning all training examples with that character to that task. For example, when $T=5$, since we only have $10$ different characters, each experience contains $2$ unique characters.

The OOD test set is produced by a distribution of samples that is not seen during training. For this, we set aside one font per character for our generalization test set, producing $10$ font/character combinations. This help us understand how well a model can compose information about the character and the font it extrapolates to unseen combinations.


\vspace{-3mm}
\subsection{Memory Replay}

For the CL experiments, we use the Avalanche library \cite{carta2023avalanche}, and the Experience Replay \cite{chaudhry2019tiny} method. Because we want to test generalization capabilities, we only use this simple method with reservoir sampling memory buffer \cite{10.1145/3147.3165}. In order to compare IID generalization versus OOD, we vary the size of the memory using 50, 100, 250, 500, and 1000. 

To run the experiments, we use a 6-layer CNN
network which we optimize using the Adam optimization algorithm \cite{adam2014KingmaB} with a learning rate of $4 \cdot 10^{-4}$ and a cross-entropy classification loss. We performed a hyper-parameter search such that we were able to replicate results from the original paper \cite{lacoste2020synbols}.

\begin{figure}[b!]
    \centering
    \vspace{-3mm}
    \includegraphics[width=0.92\linewidth]{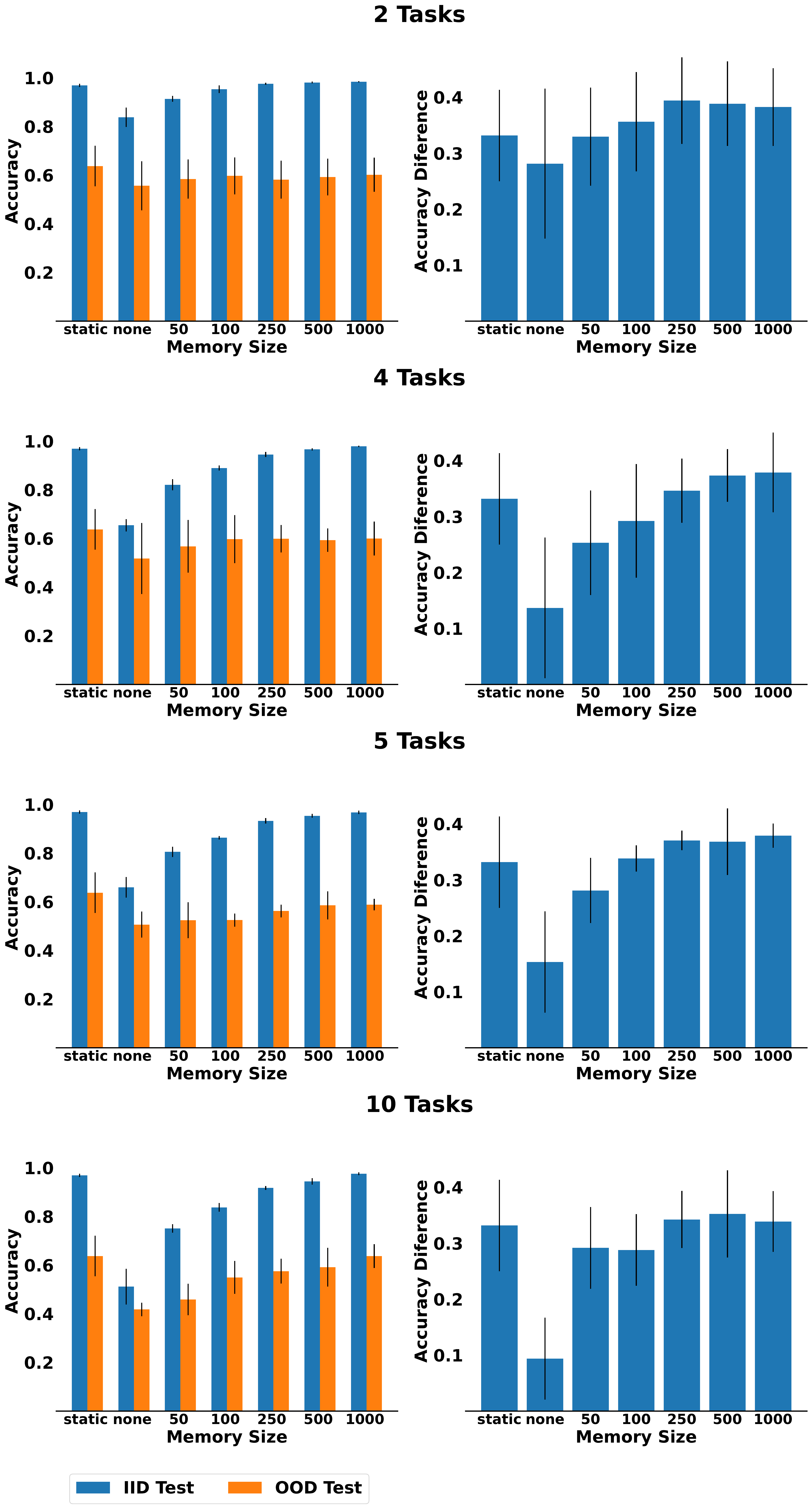}
    \vspace{-1mm}
    \caption{Bar chart for the performance (accuracy) on the IID and OOD test sets (left) and the absolute difference between both (right), for a different number of tasks (2, 4, 5, 10). Looking at the graph on the left we can see it is possible to match the performance on the IID test set to the static setting using memory, but not the generalization set. We can also notice that by increasing the memory size, the difference between both performances also increase.}
    \label{fig:baseline-generalization-gap}
    \vspace{-5mm}
\end{figure}

\vspace{-2mm}

\section{Results}

\subsection{Continual Learning Generalization}

The first thing we aim at assessing is how generalization is affected when we use a memory-based method in different CL scenarios. As a baseline, we train the model using the entire training set, which we call static training, and we can observe a gap between the accuracy achieved in the static setting and different levels of memory size in Figure \ref{fig:baseline-generalization-gap}. As expected, we need a proper representation of the training distribution to achieve similar results, otherwise, it will overfit to a subset of the training set.

If we look at the gap between IID and OOD test sets, we see that as the memory size increases the IID performance increases until it achieves similar results to the static training. However, OOD remains lower, displaying a generalization gap between both, as shown in Figure \ref{fig:baseline-generalization-gap} (right). Also, longer training sequences need larger memories to match the performance of static training. Suggesting that memory replay lacks the means to generalize to OOD data and focuses its performance to know distributions.


\vspace{-3mm}
\subsection{Representation Inspection}

\begin{figure}[t]

    \subfigure[Probe trained on IID data.]{\label{fig:iid-all-probing}
        \includegraphics[width=0.45\textwidth]{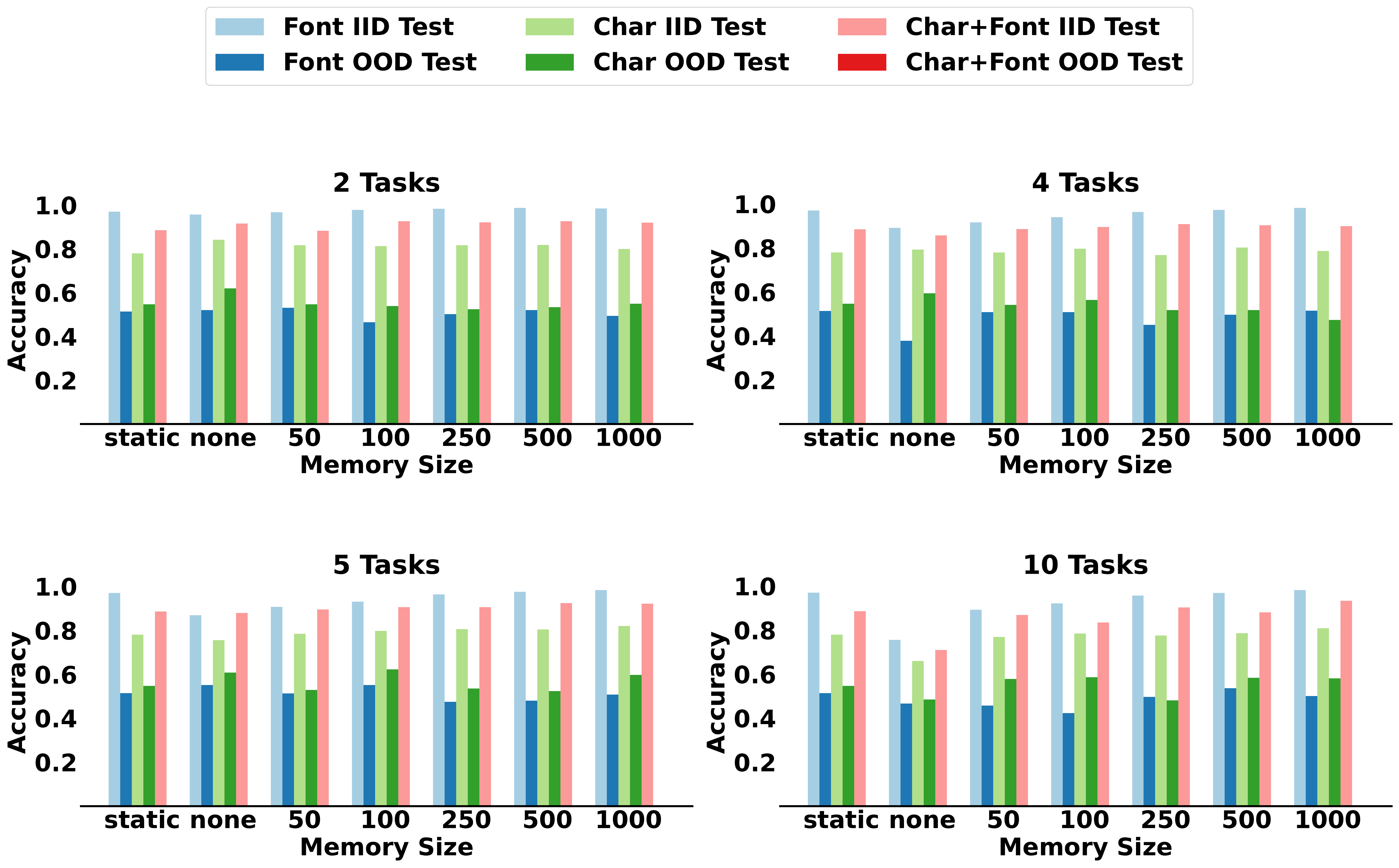}}

    \subfigure[Probe trained on IID + OOD data.]{\label{fig:ood-all-probing}
        \includegraphics[width=0.45\textwidth]{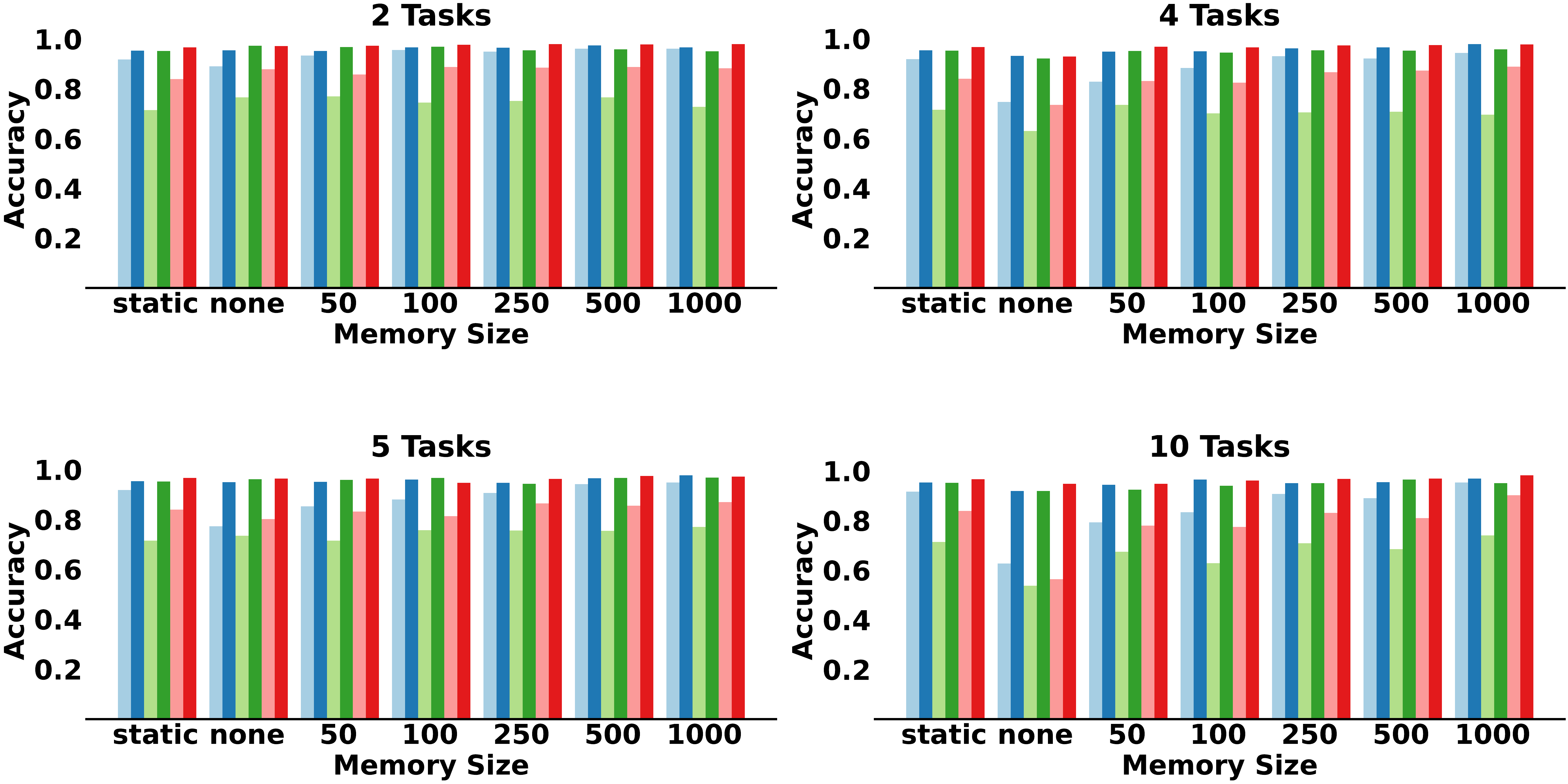}}
        

    

    \vspace{-3mm}
    \caption{Result for IID and OOD test accuracy of the font ({\color{cyan} light blue} and {\color{blue} blue} respectively), char ({\color{lime} light green} and {\color{green} green} respectively) and font/char ({\color{pink} light red} and {\color{red} red} respectively). On the left, we have results by training the probe only with the IID training set, on the right we have the results when training with IID + OOD training set.}
    \vspace{-5mm}
\end{figure}

To understand where the spurious correlations are being learned, we test the representations learned for features known to be relevant to the task. We use a linear probe, keeping the model fixed and training it to detect these features.
We show that the learned representations have the necessary information to solve the main task, since training the probe with IID and OOD data it obtains good performance in both IID and OOD test sets. This is shown in Figure \ref{fig:ood-all-probing} in light blue and blue respectively.
This suggests that is the classifier the one unable to make use of the information correctly to solve the task.
However, when training only with IID data, there is still a big gap between the IID and OOD test set, as shown in Figure \ref{fig:iid-all-probing}. This gap is similar to one presented in the previous section, and the fact that replay memory size seems not to affect this gap suggests replay is producing its effect by reducing overfitting in the IID set on the classification layer, in contrast to the representation layers of the model.


Similar to the previous experiments, we can test how much information about the characters the representation encodes, shown in Figure \ref{fig:iid-all-probing} for IID (light green) and OOD (green) test sets. The behavior is similar to the font probe, there exists a gap between IID and OOD performance when training with IID data only. 
Although smaller, a clear gap of generalization exists. 
This suggests the model is encoding information from the domain, but not confusing spuriously these two features.

\vspace{-3mm}
\subsection{Testing for Flat Modeling}

Looking at the previous results, we can see that the model produces features with the ability to solve and generalize in a compositional way, but it is not doing so. 
Our hypothesis is that the classifier is to blame for realizing these spurious relationships between character and font in such a way that there is no comprehension of compositionally between features. 
To test this, we flatten the problem and treat each font-character combination as an independent class. 
In this way, it is possible to verify if the model is actually learning to represent the combination or if it represents each concept separately so that the classifier can then combine them.

In Figure \ref{fig:iid-all-probing} and \ref{fig:ood-all-probing} we can observe that the model is able to solve the flatten task, but only when training with the corresponding combination. When training only with the original training set, the accuracy in the OOD test set is zero. However, when training with both sets, IID and OOD, the accuracy of the OOD test set is almost 100\%, light red and red respectively. 





\vspace{-2mm}

\section{Conclusion and future work}

Memory-based methods have shown high performance in various CL scenarios. However, the generalizability of these models is rarely tested. In this work, we show that these methods can only generalize in the limited context provided by the buffer only with enough memory size and that for OOD elements the performance is low. We believe it is essential to expand these types of studies to better comprehend these techniques and then propose alternatives that can generalize out-of-distribution. In future work, we seek to propose new methods that are capable of better-taking advantage of learned representations to increase their ability to generalize.

\section{Acknowledgement}

Research partly funded by National Center for Artificial Intelligence CENIA, FB210017, BASAL, ANID, and by PNRR - M4C2 - Investimento 1.3, Partenariato Esteso PE00000013 - "FAIR - Future Artificial Intelligence Research" - Spoke 1 "Human-centered AI", funded by the European Commission under the NextGeneration EU programme.

\nocite{langley00}

\bibliography{scis-workshop-template/bibliography}
\bibliographystyle{icml2023}

\newpage
\appendix
\onecolumn
\section{Appendix}

\subsection{Character representation}

In the results, we show that the model can obtain good results when training a linear probing to identify the character. this shows that the model has the information within the representation to solve the task. However, since we work in a continuous environment, it is important to know when the model obtains the relevant information to solve the task.

\begin{figure}[h]
    \centering
    \includegraphics[width=\linewidth]{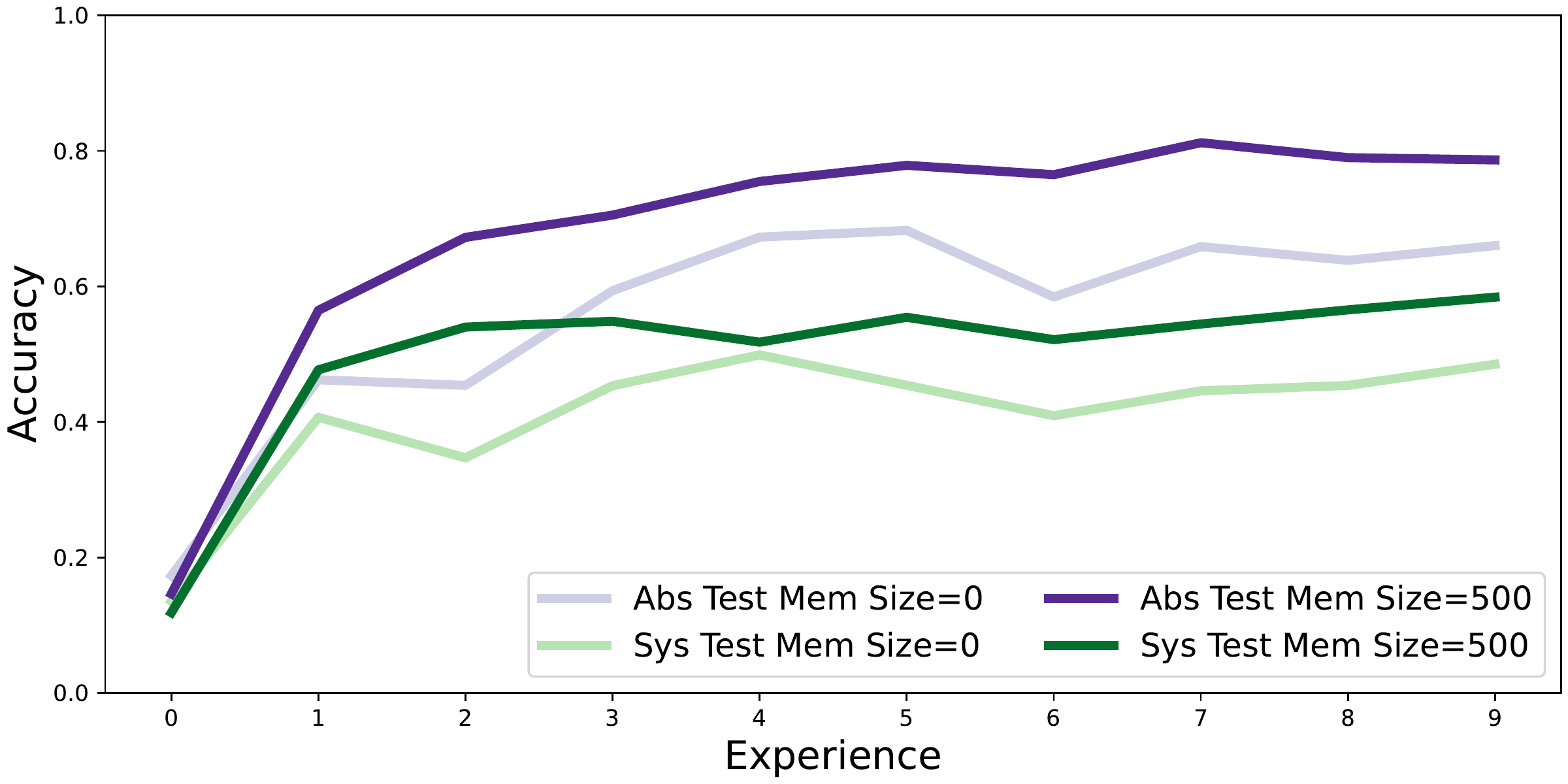}
    \caption{Character probing Accuracy after each experience for the 10 task scenario.}
    \label{fig:char-probing-exp}
\end{figure}

In Figure \ref{fig:char-probing-exp}, we can observed that the model is capable of accumulating knowledge. Even though the performance without memory is worse, it is important to note that the model is capable of good performance, both for the IID and OOD tests.

\subsection{Probing Methodology}

Using the representations at various stages of the model, we trained a linear probe to test the information stored in these.
We used a linear two-layer perceptron, the SGD with momentum training algorithm. 
We used a grid-search over the hidden layer size, $\{ 16, 32, 64, 128, 256 \}$, and learning rate $\{ 10^{-1}, 5\cdot10^{-2}, 10^{-2}, 5\cdot10^{-3} \}$, for model selection.

We use the probe to understand how much information exists in the model representation when trained in different scenarios. Figures \ref{fig:iid-all-probing} and \ref{fig:ood-all-probing} show a summary of the results. Here we show more clearly each of the results.

\begin{figure}
    \centering
    \includegraphics[width=\linewidth]{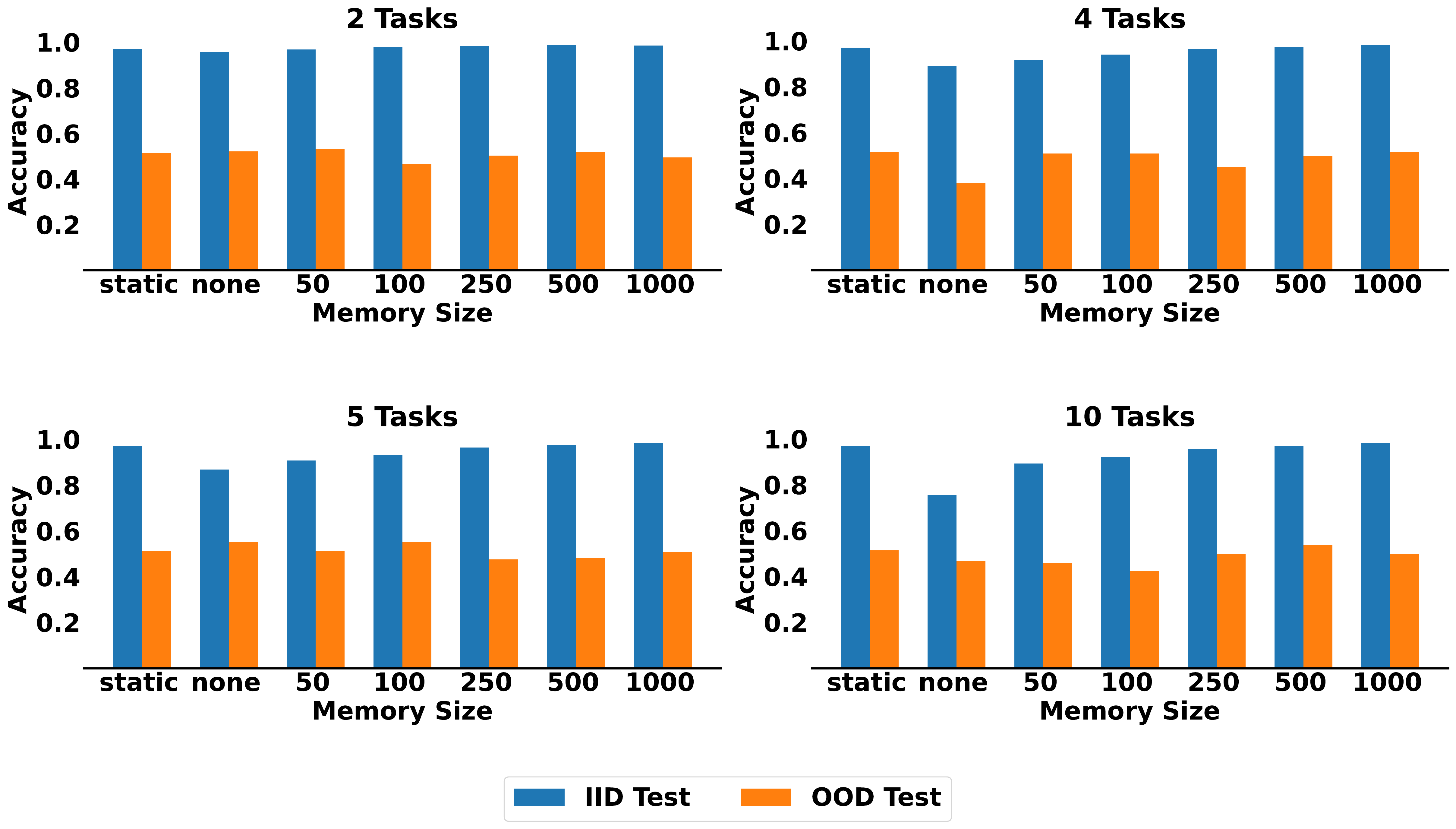}
    \caption{Result of the font probing task using the training data distribution, and testing on the IID (blue) and OOD (orange) test set.}
    \label{fig:iid-font-probing}
\end{figure}

\begin{figure}
    \centering
    \includegraphics[width=\linewidth]{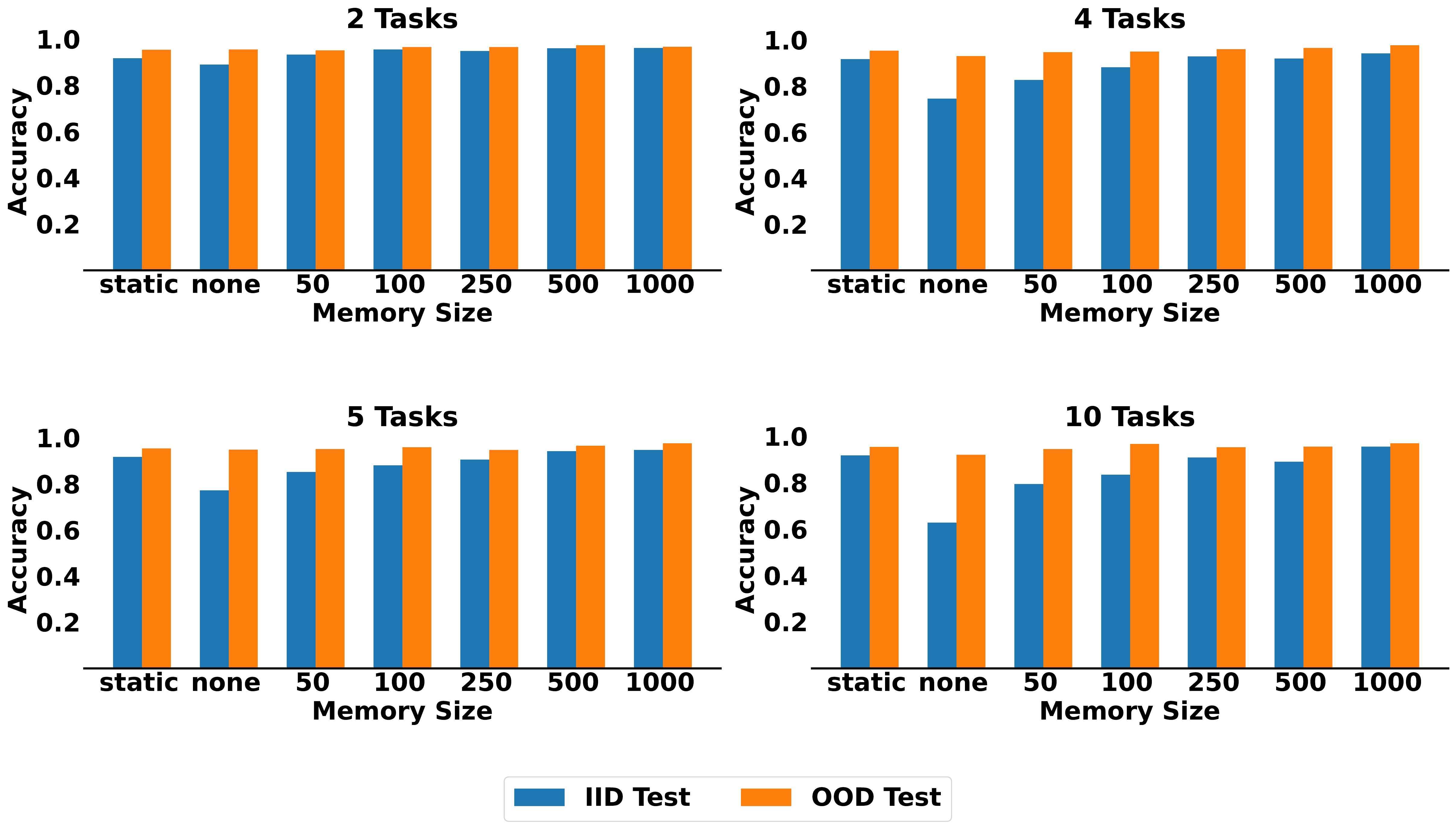}
    \caption{Result of the font probing task using the complete data distribution, IID + OOD, and testing on the IID (blue) and OOD (orange) test set.}
    \label{fig:ood-font-probing}
\end{figure}

Figures \ref{fig:iid-font-probing} and \ref{fig:ood-font-probing} show the performance of applying the probe for the font task with the training distribution, IID, and the complete distribution, IID + OOD, respectively. 

\begin{figure}
    \centering
    \includegraphics[width=\linewidth]{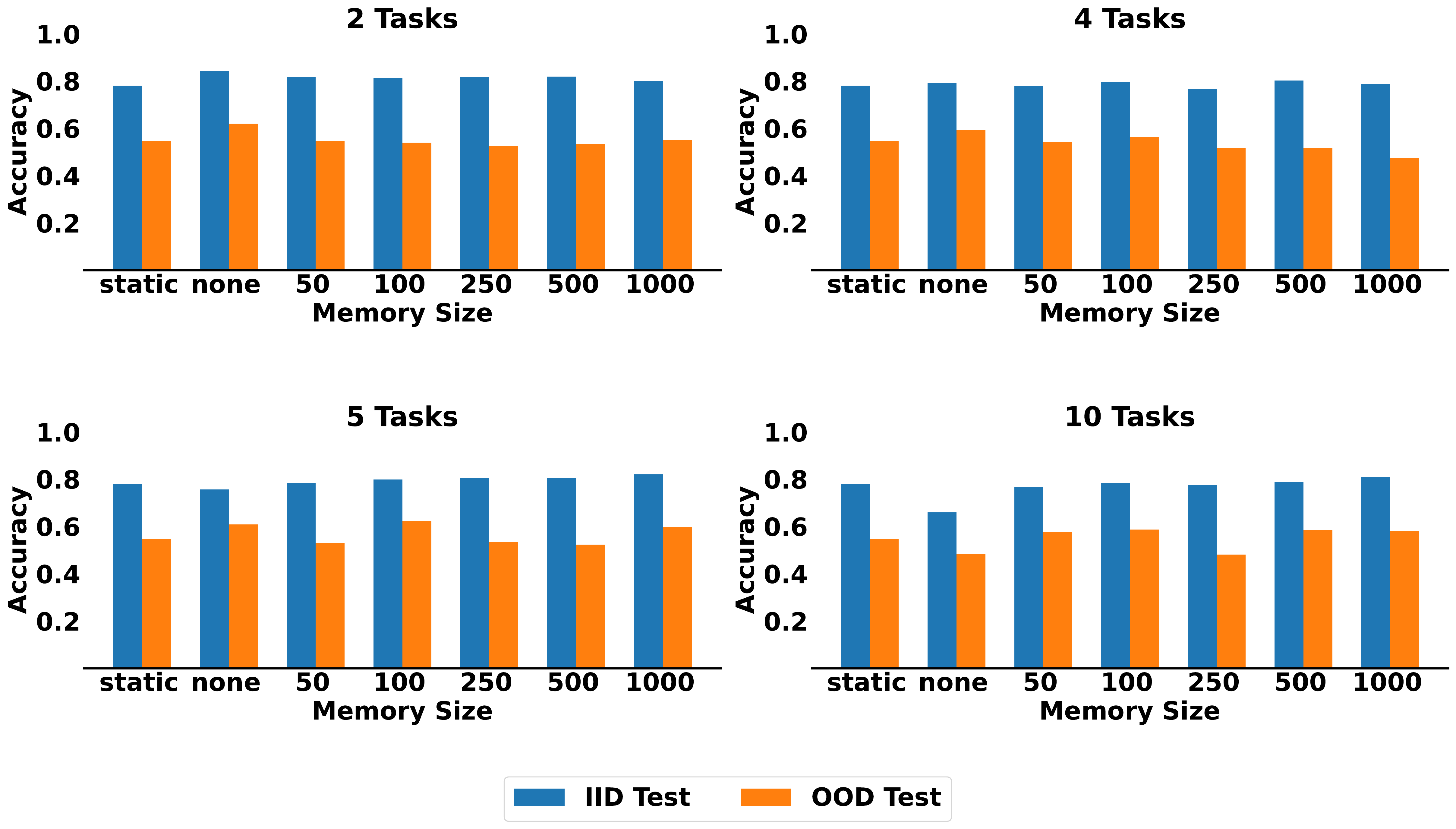}
    \caption{Result of the char probing task training with only the train distribution, IID. Testing on the IID (blue) and OOD (orange) test set.}
    \label{fig:iid-char-probing}
\end{figure}

\begin{figure}
    \centering
    \includegraphics[width=\linewidth]{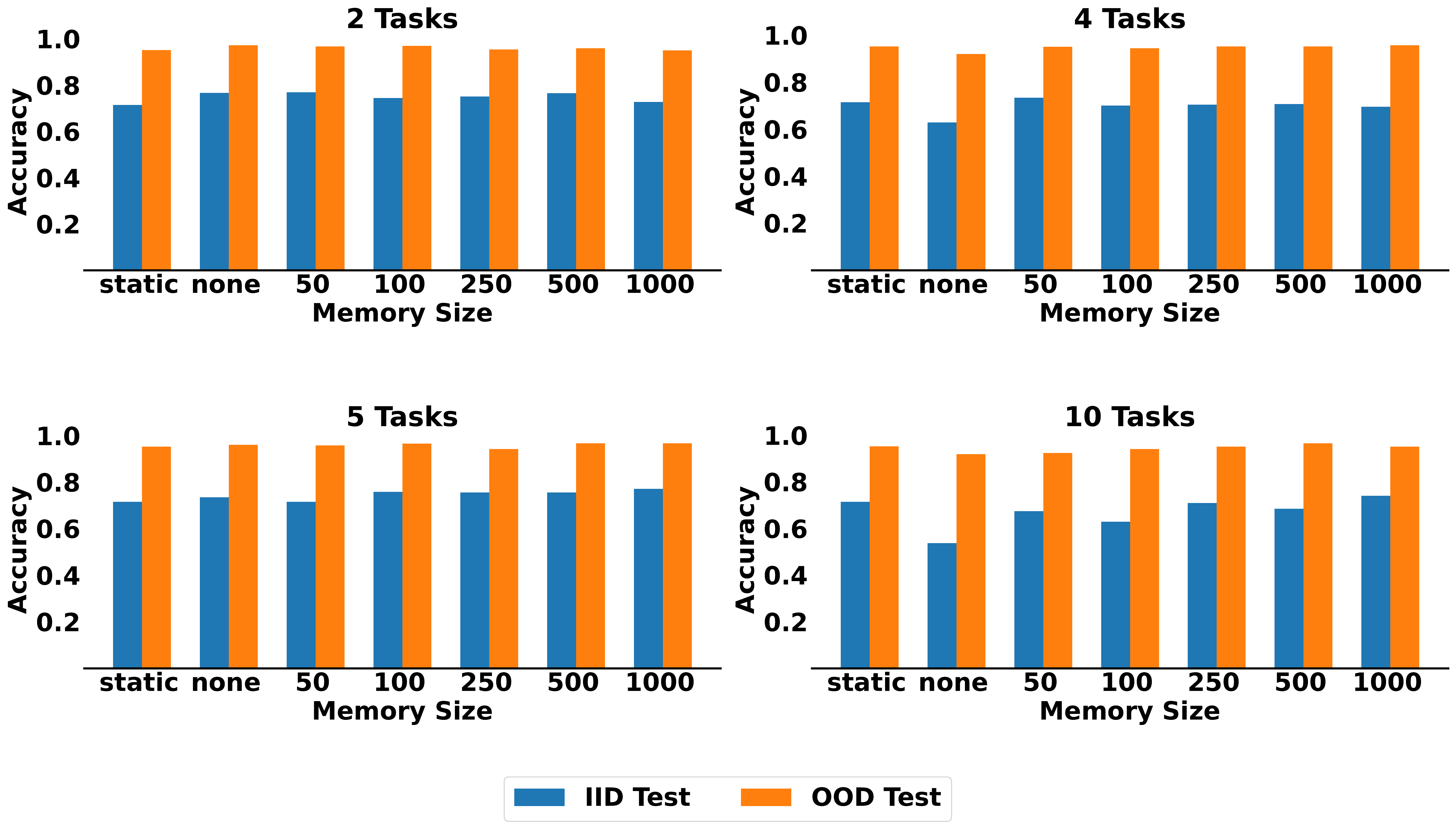}
    \caption{Result of the char probing task using the complete data distribution, IID + OOD. Testing on the IID (blue) and OOD (orange) test set.}
    \label{fig:ood-char-probing}
\end{figure}

Figures \ref{fig:iid-char-probing} and \ref{fig:ood-char-probing} shows the performance of applying probing for the character task with the training distribution, IID, and the complete distribution, IID + OOD, respectively.

\begin{figure}
    \centering
    \includegraphics[width=\linewidth]{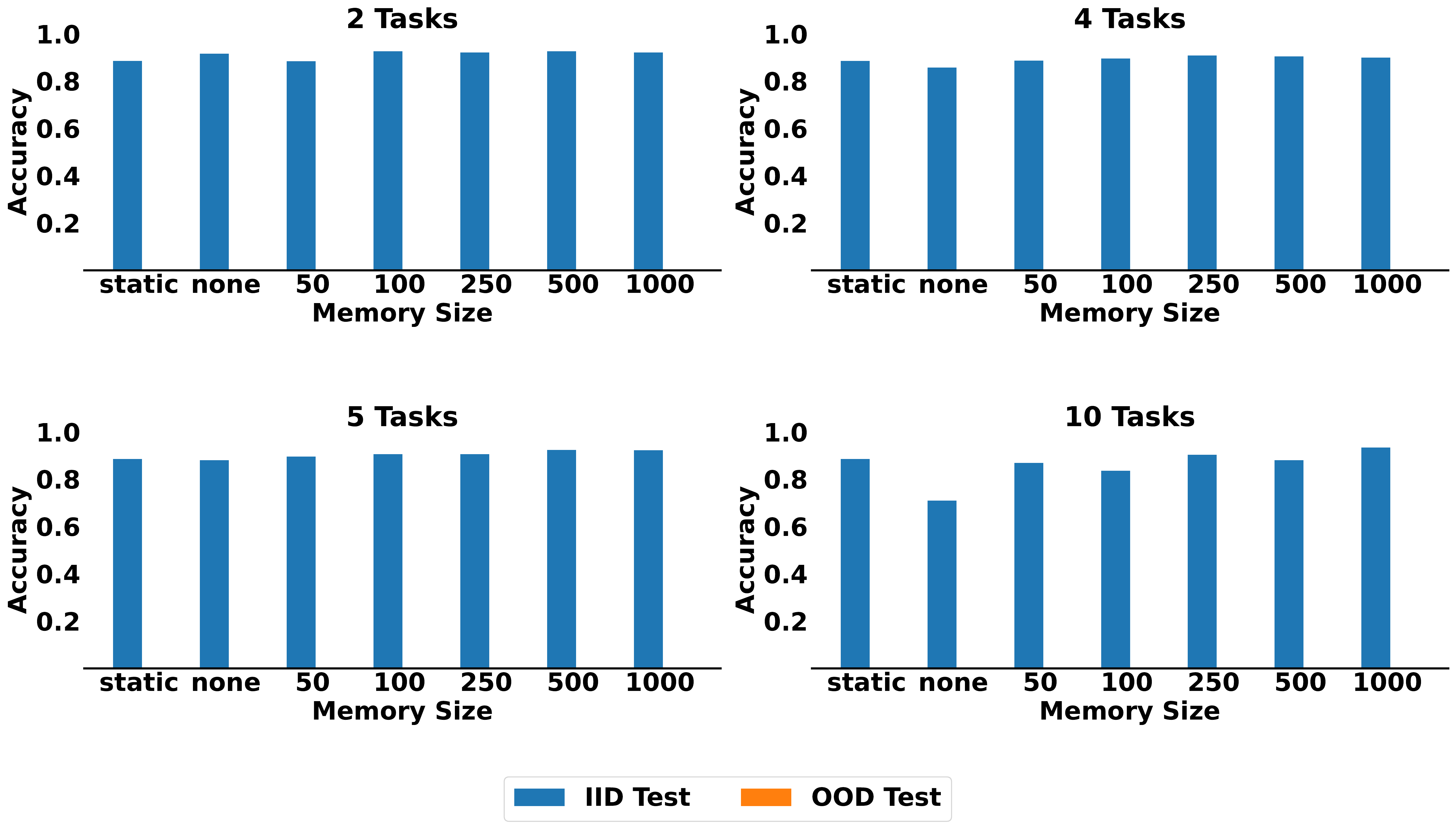}
    \caption{Result of the char/font probing task using the training set. Note that the result achieved in the OOD test are zero because the models haven't seen any data with that label during training. Similar to previous figures, IID is blue and OOD is orange.}
    \label{fig:iid-char-font-probing}
\end{figure}

\begin{figure}
    \centering
    \includegraphics[width=\linewidth]{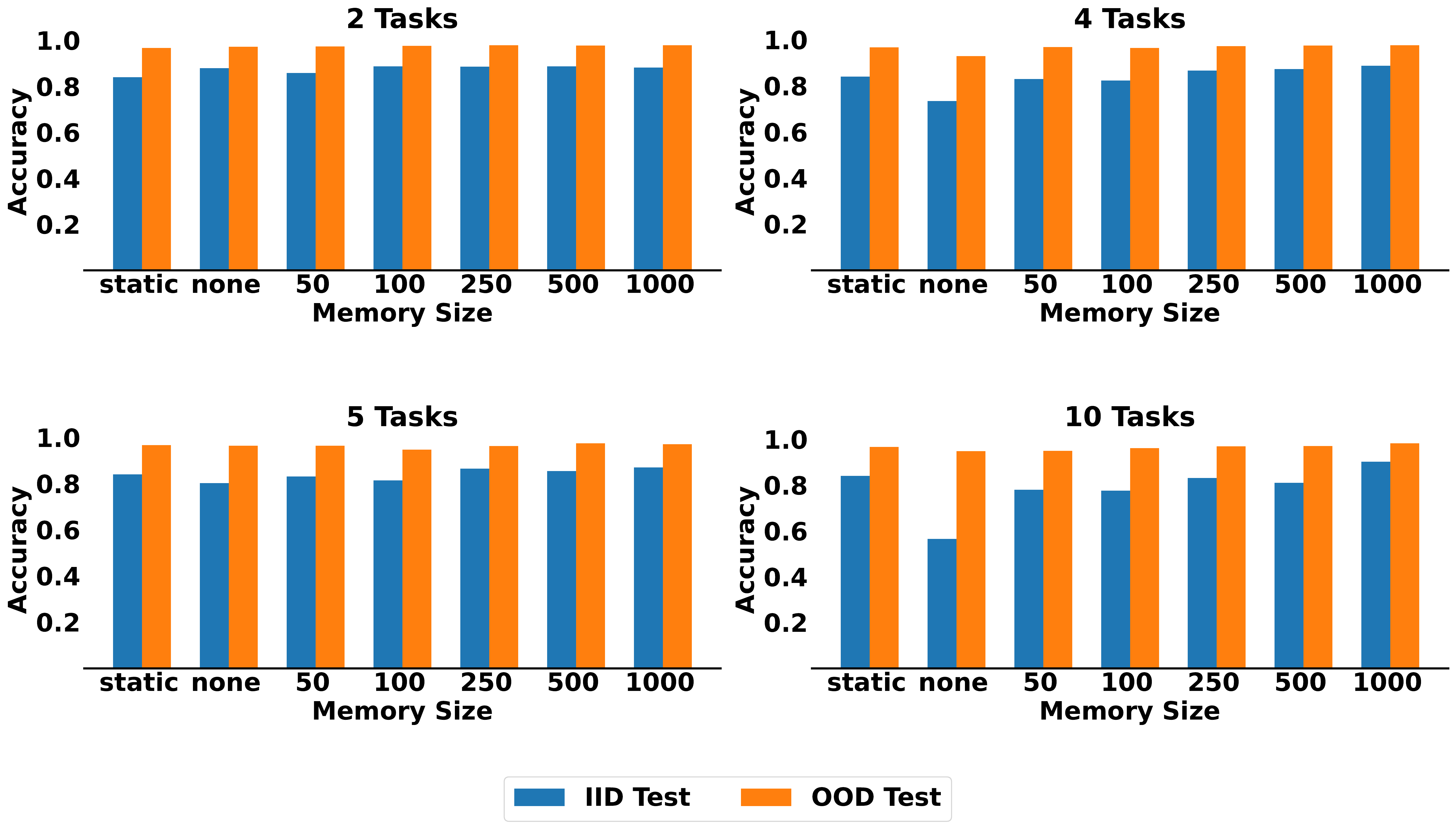}
    \caption{Result of the char/font probing task using the complete data distribution, IID + OOD. Testing on the IID (blue) and OOD (orange) test set.}
    \label{fig:ood-char-font-probing}
\end{figure}

Figures \ref{fig:iid-char-font-probing} and \ref{fig:ood-char-font-probing} show the performance of applying probing for the character task with the training distribution, IID, and the complete distribution, IID + OOD, respectively.

\end{document}